# A Physics-Informed Machine Learning Approach for Solving Heat Transfer Equation in Advanced Manufacturing and Engineering Applications

Navid Zobeiry[1*], Keith D. Humfeld[2]

* Corresponding Author, email: navidz@uw.edu

[1] Materials Science & Engineering Department, University of Washington, Seattle, WA

[2] DuoTech LLC., Federal Way, WA

**Abstract**

A physics-informed neural network is developed to solve conductive heat transfer partial differential equation (PDE), along with convective heat transfer PDEs as boundary conditions (BCs), in manufacturing and engineering applications where parts are heated in ovens. Since convective coefficients are typically unknown, current analysis approaches based on trial and error finite element (FE) simulations are slow. The loss function is defined based on errors to satisfy PDE, BCs and initial condition. An adaptive normalizing scheme is developed to reduce loss terms simultaneously. In addition, theory of heat transfer is used for feature engineering. The predictions for 1D and 2D cases are validated by comparing with FE results. It is shown that using engineered features, heat transfer beyond the training zone can be predicted. Trained model allows for fast evaluation of a range of BCs to develop feedback loops, realizing Industry 4.0 concept of active manufacturing control based on sensor data.

1. **Introduction**

From a pure mathematical point of view, in a given manufacturing process where a material undergoes several transformations (e.g. phase transformations in additive manufacturing of metallic or plastic parts, or convective curing of thermoset composites), the process can be represented using a series of governing Partial Differential Equations (PDEs) (Fernlund et al., 2018; Zobeiry et al., 2016; Zobeiry and Poursartip, 2015). These PDEs are often derived based on conservation laws such as mass, momentum, and energy conservations throughout the processing (Zobeiry and Humfeld, 2019). From an industrial perspective, it is essential to control the behaviour of the material during processing to ensure end part quality. This is often achieved by precise control of Boundary Conditions (BCs) such as pressure and temperature histories during the fabrication process. The transformations of the material subjected to these BCs is usually studied prior to processing by solving the governing PDEs using numerical methods such as Finite Elements (FE) (Johnston et al., 1996). In addition to process simulation, in process measurements using thermocouples or pressure sensors are often used in critical locations for validation of numerical models, or for feedback loops in an active control approach (Erol et al., 2016; Zobeiry et al., 2019a). However, considering the complexity of a fabrication process, controlling process parameters and precise measurement of material behaviour during processing is not a trivial task. For example, in convective heating of parts in ovens, variation of airflow and consequently variation of heat transfer coefficients (i.e. unknown or variable BCs.), lead to formation of hot and cold spots that in turn results in thermal gradients and thermal lag in the part (Fernlund et al., 2018; Park et al., 2017; Zobeiry et al., 2019a). For such a process, controlling the air temperature in the oven as a proxy to control the part temperature is quite challenging. In addition to uncertainty in fabrication process and unknown BCs, as the heat is transferred and then diffused to the center of the part via conduction, different zones in the material undergo different temperature histories. This is schematically shown in Figure 1 where a part with a thickness of $L$ is heated in a convective oven with asymmetric BCs ($h_1$ and $h_2$). As the air temperature is increased and then held at a constant temperature, the part temperature lags behind the air temperature. However, middle of the part and top and bottom surfaces undergo different temperature histories (Zobeiry et al., 2019a).

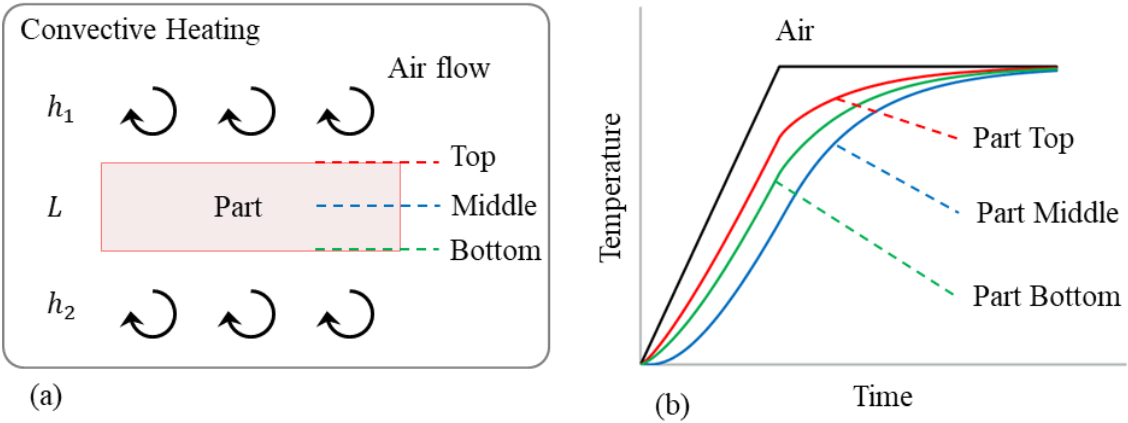

*Figure 1: (a) Convective heating of a part in an oven, and (b) Variation of temperature histories at different locations of the part.*

For such a complex problem, established industrial practices such as thermal profiling combined with process simulation using FE models are frequently employed by engineers for thermal management. However, FE models are not suitable to consider uncertainties in manufacturing processes including unknown BCs. As a result, engineers often rely on trial and error simulations based on different assumptions for BCs during processing. This combines with the fact that high fidelity FE tools are inherently slow, makes such an approach impractical for an industry 4.0 setting where fast and near real-time simulation capabilities are needed (Agrawal and Choudhary, 2016; Erol et al., 2016).

The rise of Machine Learning (ML) and AI in recent years offer an opportunity to develop fast surrogate ML models to replace traditional FE tools in manufacturing and general engineering applications. Several approaches have been explored in the literature in recent years:

- Theory agnostic ML models trained using physical data: Using limited physical data obtained from experiments or embedded sensors during manufacturing, predictive ML models have been developed for different applications including turbulence modeling, and guiding assembly of aerospace parts (Manohar et al., 2018; Singh et al., 2017). The main challenge in these applications is developing ML techniques suitable for small physical datasets.
- Theory agnostic ML models trained using numerical data: by automating FE models to generate large numerical datasets, surrogate ML models have been trained in many applications including turbulence modeling (Wang et al., 2017), simulation of manufacturing processes and failure in advanced composites (Zobeiry et al., 2020b, 2020a, 2019b),  and stress analysis (Liang et al., 2018). One main challenge is the trade-off between fidelity and speed in FE tools, and consequently accuracy of the trained ML model. High fidelity FE tools are slow which limits their applicability to generate large numerical datasets. On the other hand, low fidelity FE tools are typically much faster and can be used to generate larger datasets. This generally comes at the cost of reduction in the accuracy of the model.

- Theory-guided ML models trained using a combination of governing physical laws, numerical data and physical data: It has been shown that by combining several data streams, and designing architecture and features of ML models based on governing physical laws, some limitations of previous approaches can be addressed (Karpatne et al., 2017; Wagner and Rondinelli, 2016; Zobeiry et al., 2020a). This includes using small physical datasets and low-fidelity FE tools to train ML models with acceptable accuracies even beyond their training zones (Zobeiry et al., 2020b, 2019b).
- Physics-informed ML models trained using governing PDEs: Instead of solving PDEs using FE tools to generate training datasets, in several studies governing PDEs have been used directly to train ML models (Han et al., 2018; Raissi, 2018; Raissi et al., 2019, 2017a, 2017b; Sirignano and Spiliopoulos, 2018). In these studies, Physics-informed Neural Networks (PINNs) are typically trained by defining the loss function as the error in satisfying the governing PDE. By reducing the loss function, PINN is trained to ultimately satisfy the PDE. One key advantage of such an approach compared to previous approaches is that pre-generated training data such as FE results is not needed. In addition, once trained using BCs as inputs for the PINN, computational time will be much faster than FE models. This makes this approach quite suitable for implementation in industry 4.0 applications.

Following previous studies and established approaches to solve PDEs using ML (Han et al., 2018; Raissi, 2018; Raissi et al., 2019, 2017a, 2017b; Sirignano and Spiliopoulos, 2018), a PINN is developed in this study to solve the heat transfer PDE with convective BCs in a representative manufacturing setting. In addition, physics-informed features are engineered based on the theory of heat transfer to accurately represent the underlying physics using the trained PINN. It is shown that using this approach, a PINN can predict heat transfer even outside its training zone. Comparison with FE results is used to validate the approach in 1D and 2D cases. The developed PINN allows for fast evaluation of heat transfer problems with various convective BCs. In an industry 4.0 setting, this leads to development of near real-time feedback control loops to adjust the process parameters and control the temperature history of the part.

## 2. Method

The general heat transfer equation for a given part can be written as (Incropera et al., 2007):

$$\frac{\partial}{\partial t}(\rho C_P T) = \frac{\partial}{\partial x}\left(k_{xx}\frac{\partial T}{\partial x}\right) + \frac{\partial}{\partial y}\left(k_{yy}\frac{\partial T}{\partial y}\right) + \frac{\partial}{\partial z}\left(k_{zz}\frac{\partial T}{\partial z}\right) + \dot{Q} \qquad (1)$$

In which $T$ is the temperature, $\rho$ is the part density, $C_p$ is the part specific heat capacity, $k$ is the part conductivity and $\dot{Q}$ is the rate of heat generation in the part. For sake of simplicity, the method is explained here for one-dimensional heat transfer (i.e. 1D). But this can be easily extended to higher dimensions which will be discussed later. The general heat equation can be simplified to represent the case of the one-dimensional heat transfer with no heat generation term:

$$\frac{\partial T}{\partial t} - \alpha\frac{\partial^2 T}{\partial x^2} = 0 \qquad (2)$$

Where $\alpha$ is the thermal diffusivity defined as:

$$\alpha = \frac{k}{\rho C_P} \qquad (3)$$

The convective boundary condition is written as:

$$h(T_\infty - T_{boundary}) = k \frac{\partial T}{\partial x}\bigg|_{boundary} \qquad (4)$$

Where $h$ is the convective Heat Transfer Coefficient (i.e. BC), $T_\infty$ is the air temperature around the part and $T_{boundary}$ is the part temperature at its surface.

Suppose that the prediction made by a neural network, $f(x, t, h_1, h_2)$, is intended to be a solution to the one-dimensional heat equation for any given boundary condition. The solution's adherence to the heat transfer PDE at any given point can be quantified as:

$$Error_{PDE} = \alpha \frac{\partial^2 f(x, t, h_1, h_2)}{\partial x^2} - \frac{\partial f(x, t, h_1, h_2)}{\partial t} \qquad (5)$$

If the prediction made by the neural network is a perfectly trained solution to the one-dimensional heat equation, this error term will be zero at every point. For the two boundaries located at $x_1$ and $x_2$, the adherence to the convective boundary condition differential equations can be quantified as:

$$Error_{BC1} = -(T_\infty(t) - f(x_1, t, h_1, h_2)) + \frac{k}{h_1} \frac{\partial f(x, t, h_1, h_2)}{\partial x}\bigg|_{x=x_1} \qquad (6)$$

$$Error_{BC2} = (T_\infty(t) - f(x_2, t, h_1, h_2)) - \frac{k}{h_2} \frac{\partial f(x, t, h_1, h_2)}{\partial x}\bigg|_{x=x_2}$$

If the prediction made by the neural network is a perfect solution to boundary conditions, these error terms will be zero at any given time and for any given heat transfer coefficient. Heat transfer problems are frequently solved with an Initial Condition (IC), such as the one-dimensional solid being in thermal equilibrium at the beginning of the problem. One productive way to think of an IC is by recognizing that the IC simply describes a boundary condition applied at the time-dimension boundary $t = 0$. If the neural network intended to also represent a solution to such a boundary condition, the adherence to that boundary condition can be quantified as:

$$Error_{BC0} = T_\infty(0) - f(x, 0, h_1, h_2) \tag{7}$$

For a collection of training data points, the loss term for training the neural network can be defined based on Equations 5-7:

$$Loss = Loss_{PDE} + \lambda_0 Loss_{BC0} + \lambda_1 Loss_{BC1} + \lambda_2 Loss_{BC2} \rightarrow \tag{8}$$

$$Loss = \frac{1}{N_{PDE}} \sum_{i=1}^{N_{PDE}} Error_{PDE}^2 + \frac{\lambda_0}{N_{BC0}} \sum_{i=1}^{N_{BC0}} Error_{BC0}^2 + \frac{\lambda_1}{N_{PDE}} \sum_{i=1}^{N_{BC1}} Error_{BC1}^2 + \frac{\lambda_2}{N_{BC2}} \sum_{i=1}^{N_{BC2}} Error_{BC2}^2$$

Where $\lambda$ values are scaling factors to normalize loss terms. Each term of the loss function is designed to calculate the mean square of the error term over the points for which that error term was evaluated. For a neural network that perfectly represents the solution to the one-dimensional heat equation, all of the loss values, evaluated over any arbitrary sets of points will sum to zero. Combining multiple loss functions such as these four loss functions into one cumulative loss function, however, faces a recognized challenge that the relative magnitudes of the losses impact the model training and result. If the magnitude of one loss function is significantly greater than the magnitude of the others, or if the sensitivity of one loss function to a change in weightings is significantly greater than the sensitivity of the other loss functions to a change in weightings, the neural network will train to a solution that minimizes one of the loss functions but is not apparently influenced by the other loss functions. This issue is addressed later in the implantation section by developing an adaptive normalizing scheme.

## 3. Implementation and Training of a Physics-Informed Neural Network

Training of a PINN to solve the heat transfer PDE with convective BCs was implemented in Python (V3.6.8), using Tensorflow and Keras libraries (V2.10). In this section, we discuss details of implementations including PINN architecture, theory-guided feature engineering, choice of activation function and adaptive normalization scheme. The training of PINN was based on selecting random $(x, t, h)$ batches in each epoch and minimizing the loss function (Equation 8) using built-in Keras optimizers to obtain weights and biases of the neural network to satisfy heat transfer PDE. Based on performing a grid search, Adam optimizer with a learning rate of 0.0001 was used with a batch size of 150 for each loss term. In most cases, an 100K epochs were used for training.

### 3.1. *Physics-informed NN Architecture and Engineered Features*

Potentially, any neural network with a dense architecture can be trained with previously defined loss terms to represent the solution to the heat transfer PDE. An example of such a network is schematically shown in Figure 2. In addition to implementing physics-informed loss function, physics-informed feature engineering can be used as a complementary approach to better represent the governing physics. In the absence of heat generation and convection, the heat equation can be solved analytically by separation of variables (Incropera et al., 2007):

$$T(x,t) = T_0 + (T_{Max} - T_0) \sum_n A_n exp\left[-\frac{k}{\rho c_p}\left(\frac{n\pi}{L}\right)^2 t\right] cos\left(\frac{n\pi}{L}x\right) \tag{9}$$

Where $A_n$ are weights and $n$ is an integer. In the presence of heat generation or convection, an analytic solution would take a different form if the problem were analytically solvable. That said, some of the underlying physics could be captured by engineering the features of the neural network to conform to the analytic form of the solution of the heat equation without heat generation or convection.

To implement feature engineering, a first layer was made for the neural network by combining two pre-layers of terms as shown in Figure 3. The first of the pre-layers took the position argument as an input, applied a trainable weighting and bias, and applied a sine function activation function; in the presence of a trainable bias, the sine and cosine functions are interchangeable. The second of the pre-layers took the time argument as an input, applied a trainable weighting and bias, and applied an exponential activation function. These two pre-layers were then multiplied on a 1:1 basis to create a layer with a number of terms of the form

$$exp[at + a_0]sin(bx + b_0) \qquad (10)$$

Using this method of construction, with trainable weightings on the two pre-layers, the training of the neural network has responsibility to find weightings that give the proper form of the solution. The architectures shown in Figure 2 and Figure 3 are referred to as NN and PINN respectively in this study. The performance of these two architectures will be compared in the validation section. Using a grid search, 6 hidden layers with 32 nodes per layer was selected in both cases. In addition, 32 engineered features based on Equation 10 were used in the PINN.

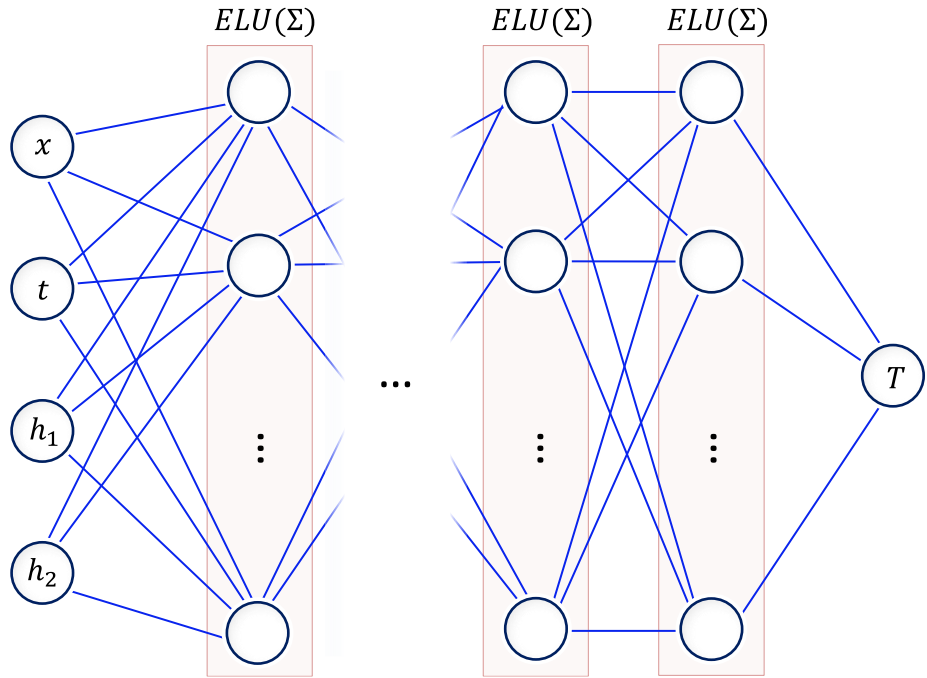

Figure 2: Schematic of a neural network with a dense architecture to solve the heat transfer PDE. This is referred to as NN in this study.

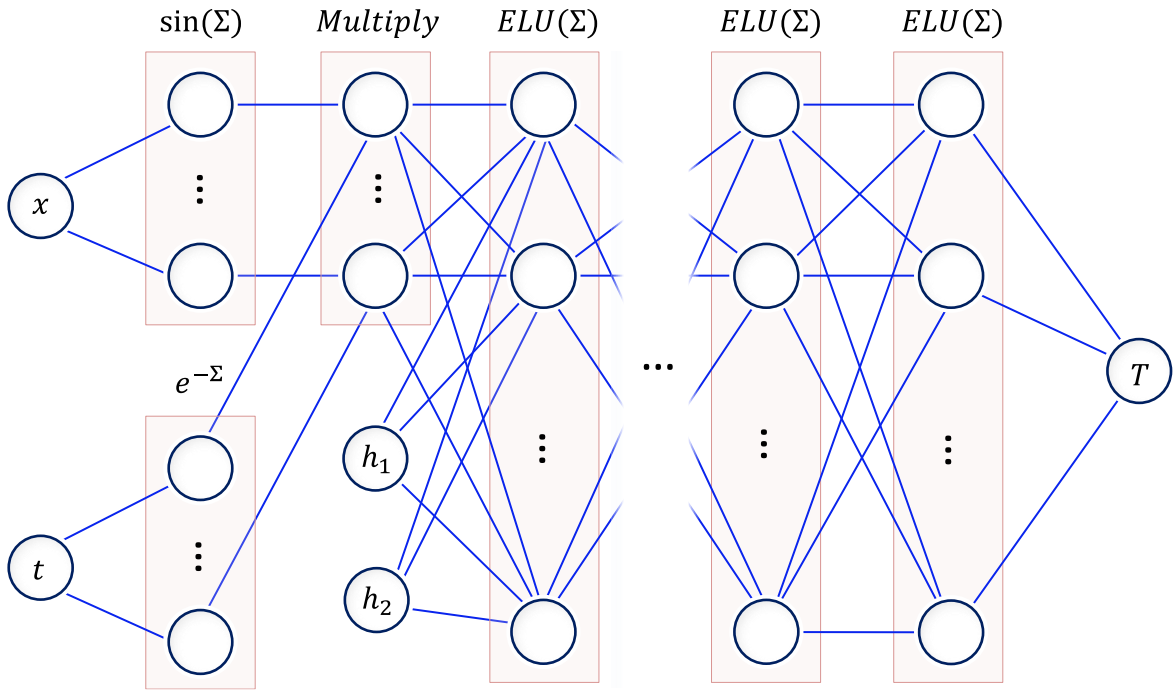

Figure 3: Schematic of a neural network with physics-infirmed engineered features to solve the heat transfer PDE. This is referred to as PINN in this study.

## 3.2. *Activation Function*

Choice of activation function has a significant impact on the success of training PINNs to solve PDEs. Common activation functions such as sigmoid or *tanh* may suffer from vanishing gradients problem. During training following the backpropagation algorithm, the derivative of the loss function with respect to the weights of each layer must be calculated. In this calculation, the derivative of activation functions is multiplied by itself several times, equal to the layer distance from the output. For an activation function where the derivative is squeezed into a narrow range, training of a NN with many hidden layers may not be successful. ReLU function on the other hand does not suffer from this problem and as a result widely used in many deep learning applications. Several activation functions and their derivates are compared in Figure 4.

Training a PINN to solve a PDE using ReLU activation function poses challenges. For solving a PDE such as heat transfer, it is necessary to take the first and second derivatives of the NN with respect to network inputs to calculate the loss function. As a result, training which is based on calculation of the derivate of the loss function, includes second and third derivates of the activation function. However, higher order derivatives of ReLU are equal to zero as shown in Figure 4 for the second derivative. This makes the training process ineffective:

$$\frac{d^2 ReLU(\Sigma)}{dx\Sigma^2} = 0 \tag{11}$$

As a result, it is necessary to train a PINN with an activation function that has a nonzero second derivative and still allows effective deep network training. For this, we propose the ELU activation function for training PINNs (Figure 4):

$$ELU(\Sigma) = \begin{cases} \Sigma & \Sigma \geq 0 \\ e^{\Sigma} - 1 & \Sigma < 0 \end{cases} \qquad (12)$$

Using ELU activation functions with 6 hidden layers, this PINN was trained to accurately solve the heat transfer PDE. Trials with *tanh* function on the other hand, were not successful with a range of 4-10 hidden layers.

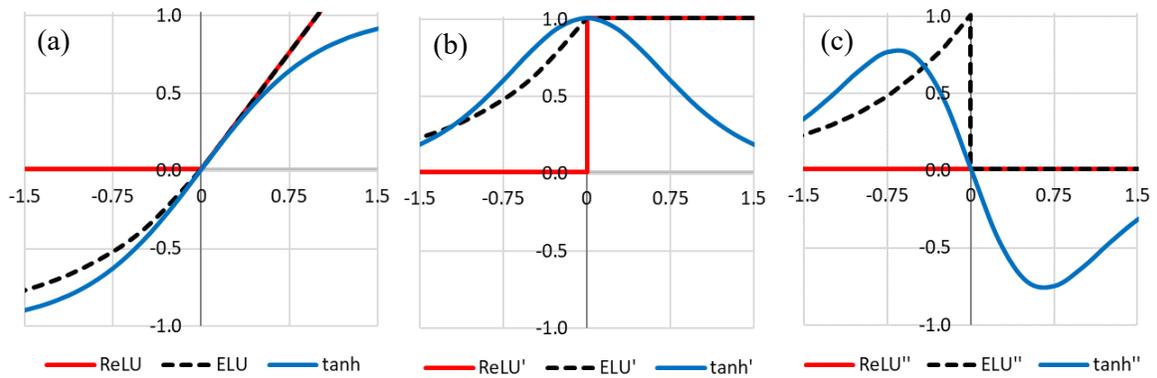

*Figure 4: (a) ReLU, ELU and tanh activation functions, (b) derivatives of activation functions, and (c) second derivatives of activation functions*

### 3.3. Adaptive Normalization Factors

The issue with relative magnitudes of loss values as described previously was resolved by using an adaptive normalization scheme. At the beginning of the training, the model is initialized using the Glorot uniform initializer in Keras (i.e. uniform distributions for weights and biases). Since this does not represent the solution to the system of equations, the relative magnitudes of the loss terms in Equation 8 could be very different. This means that the largest error terms may dominate the training, such that error terms with smaller magnitudes would not be reduced as the training progresses. For a well-trained model however, loss terms are defined such that the individual error terms are comparable.

The adaptive scheme was developed to update the normalization factors in Equation 8 at regular intervals, e.g. every 100 epochs. New normalization factors were determined by evaluating the ratio between the specific loss term and the greatest loss term. If the ratio of loss terms was greater than some threshold (e.g. 0.01) then the normalization factor for the specific loss term was set to unity. If the ratio of loss terms was below that threshold, i.e. the loss term was much smaller than the greatest loss term, the normalization factor was set to the loss ratio divided by the threshold. This brought the relative error term to within the threshold for the current epoch.

As training continued, if the greatest loss term trained less slowly than this specific loss term, then the normalization factor would be intensified at the next update interval. This would put the model into a scenario where the model trains towards a solution for the maximum magnitude loss term with somewhat little influence from the other loss terms. While using this adaptive normalization scheme, however, this was not problematic: as the model continued to train, this maximum magnitude loss term was decreased faster than the other terms. The interplay between some loss terms training more quickly due to being more simple and having their normalization factors decrease rapidly, and other loss terms training more quickly due to the normalized loss term being significantly larger, resulted in normalization factors gradually approaching the threshold and finally being set to unity. During some instances of model training, the adaptive scheme would set a normalization factor to unity in one update and then the loss ratio would drop down below the threshold during a later update, but with enough updates, all loss terms approach the same order of magnitude and all normalization factors became unity.

### 3.4. *Prediction Accuracy around Boundary Condition Kinks*

One of the difficulties observed in training neural networks is inaccurate predictions around kinks and discontinuities in the input data, for example in locations where $dT_{air}(t)/dt$ is not continuous (as shown in Figure 1). Such discontinuity typically results in low training accuracy in the vicinity of that kink. As discussed above, one of the key advantages of using the differential equation and boundary conditions to train a neural network, was that no pre-generated training data needed to be defined, and therefore the specific points at which the model was trained did not need to be defined prior to training. This feature was exploited to improve the accuracy in the vicinity of the external air temperature kink, by increasing the density of training points in each training batch chosen in the vicinity of the external air temperature kink. Likewise, the initial condition representing thermal equilibrium implies that there is a discontinuity in the derivative of the air temperature at the beginning of the problem. This led to inaccuracies in the predictions of the neural network for the lowest values of time; the accuracy was improved by increasing the density of training points near $t = 0$.

## 4. 1D Validation

The accuracy of the Physics-Informed Neural Network was investigated by comparing the predictions against calculations from a finite element model. The training and prediction accuracy of the PINN (Figure 3) was also compared to the training and prediction of a simple NN (Figure 2). Finally, to illustrate and validate the ability of the PINN to utilize the heat transfer coefficients on the boundaries as inputs, the trained PINN was run for a variety of heat transfer coefficients and compared to the results of a series of matching finite element analyses. In all trainings and simulations, a carbon-fiber epoxy composite part with following nominal properties was used (Zobeiry et al., 2019a):

- $k = 0.47\ W/mK$
- $\rho = 1573\ kg/m^3$
- $C_p = 967\ J/kgK$

An in-house developed FE code in python was used to solve the heat transfer PDE with convective BCs. Upon performing mesh size sensitivity, the geometry was discretized into 10 elements in each directions, and time was discretized into 5 second intervals.

### 4.1. *PINN versus FE*

The Physics-Informed Neural Network described in Section 3 was trained to predict the time-dependent solution to the 1D heat equation in a composite part of 10 mm thickness, in the scenario with a uniform initial temperature boundary condition ($T_0$=0 °C), asymmetric convective heat transfer boundary conditions ($h_1$=100 W/m²K, $h_2$=50 W/m²K), and the specific external air temperature profile illustrated in Figure 5 (a) (a heating rate of 5 °C/min to 50 °C, and then held for 5 minutes). The prediction of the neural network for the temperature history in the middle of the part is shown by the solid blue line in Figure 5 (a). To validate the PINN predictions, a finite element simulation was run for an identical problem. The temperature in the middle of the part predicted by the finite element analysis is depicted by the dotted red line in Figure 5 (a). The prediction of the neural network matches the calculation of the finite element analysis quite well with a maximum deviation less than 0.97 °C at any given time. Figure 5 (b) illustrates the equivalence of the PINN predictions and the FE simulation results 10 minutes into the simulation, by showing the temperature through the thickness of the part. The predictions are asymmetric since heat transfer coefficients are different below and above the part. The finite element analysis, performed with ten elements through the thickness, can predict temperatures only at the nodes of the elements and must interpolate between those nodes, where the PINN can be evaluated at any point along the continuum, as the x-position of every point at which the PINN was trained was selected randomly from the continuum at each epoch.

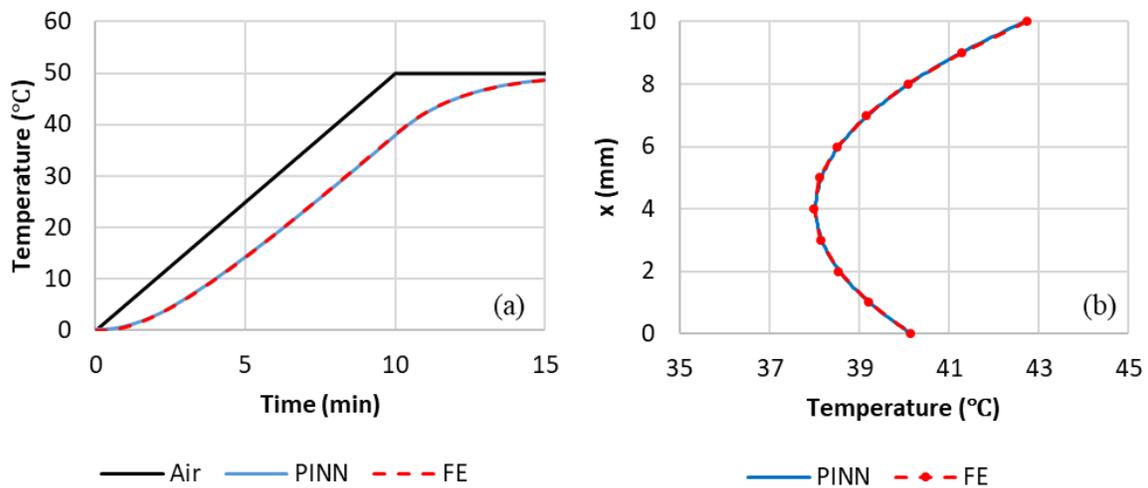

*Figure 5: (a) Temperature histories in the middle of a composite part subjected to an air temperature cycle, predicted using PINN and FE, and (b) Through-thickness temperature distribution after 10 minutes as predicted by PINN and FE.*

### 4.2. PINN versus NN

Heat transfer in a 20 mm composite part with asymmetric convective BCs ($h_1$=100 W/m²K, $h_2$=50 W/m²K) and a 15 minutes temperature cycle identical to previous case as depicted in Figure 5 (a) was modeled using three methods FE, NN and PINN. A neural network without feature engineering as depicted in Figure 2 (NN), and a neural network with feature engineering as depicted in Figure 3 (PINN), were trained for the 15 minutes temperature cycle. However, after training, they were used to predict the part temperature history 15 minutes beyond the training cycle for a total period of 30 minutes. The NN trained faster than PINN as shown in Figure 6. This illustrates that the NN reached a stable and equivalent total loss after fewer epochs than did the PINN. As shown in Figure 7, both models match finite element predictions over the entire range of the training data. Beyond the range of the training data, i.e. for times greater than 15 minutes shown in Figure 7, the PINN made predictions for considerably closer to the finite element results, while NN diverged from FE predictions. This is due to the implementation of physics-informed features in PINN to capture the underlying behavior of the heat transfer problem.

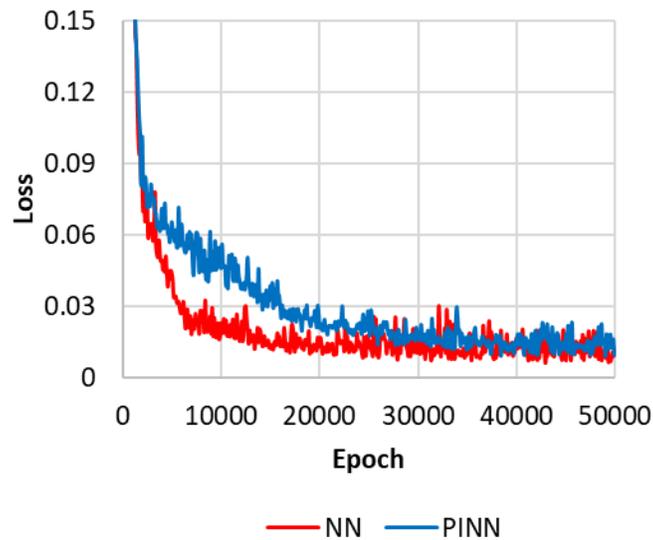

*Figure 6: Loss functions during training of NN and PINN models as shown in Figure 2 and Figure 3 respectively.*

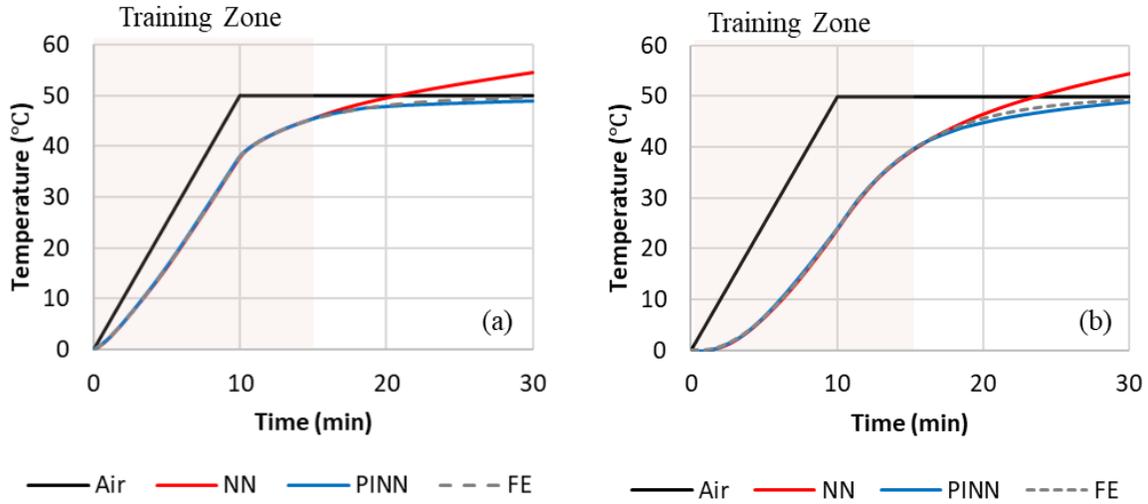

*Figure 7: (a) Temperature histories at the top surface of a part as predicted by FE, NN and PINN, and (b) Temperature histories at the middle of a part as predicted by FE, NN and PINN.*

4.3. *PINN with Convective BCs as Inputs*

Using a PINN to predict the solution to the heat equation in the presence of convective heat transfer boundary conditions was made advantageous, over using the finite element method, by extending the inputs of the neural network to include the heat transfer coefficients on both sides of the one-dimensional solid, as depicted inFigure 3. Increasing the dimensionality of the inputs of the neural network from 2 to 4 increased the number of points required to train the neural network, but this simply increased the number of epochs and the batch size required to train; a key advantage to using the physics informed neural network structure that calculates losses based on the adherence to the differential equations was that no training data needed to be generated prior to training the model. The PINN that included heat transfer coefficients as inputs was trained to similar accuracy as the two-input neural network. This extended neural network enabled predictions of the temperature as a function of time, position, heat transfer coefficient above and heat transfer coefficient below.

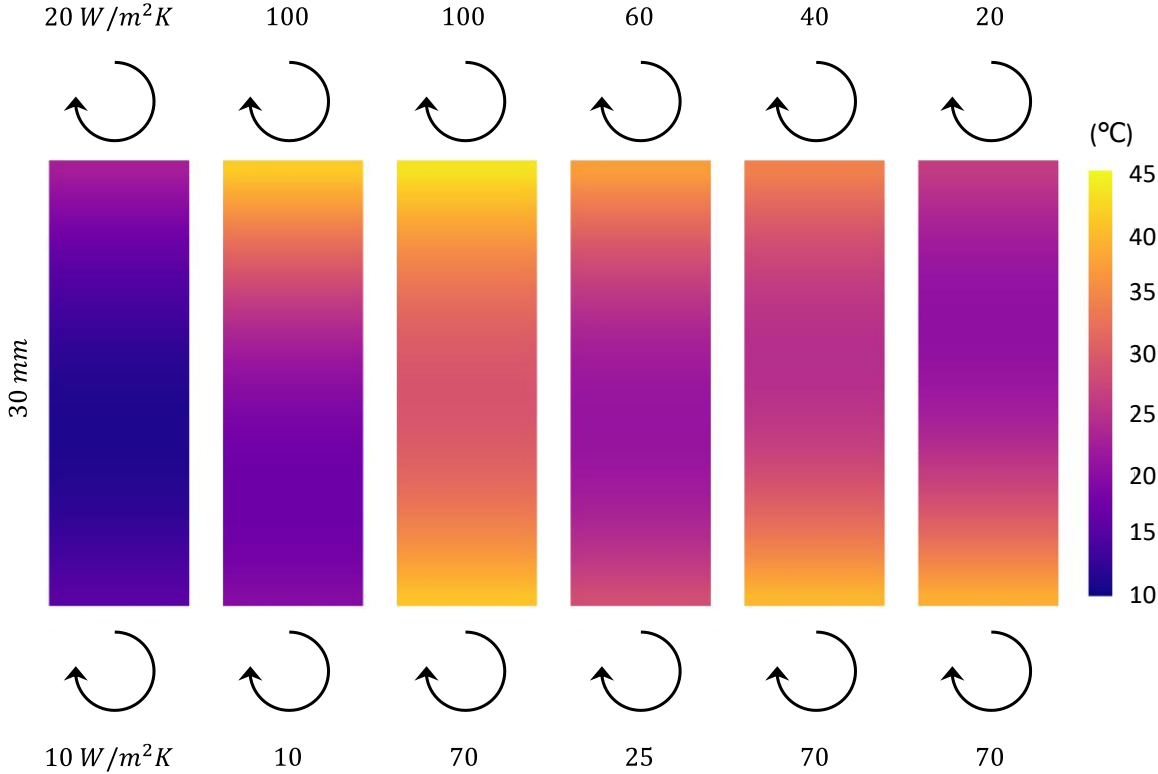

*Figure 8: Near real-time predictions of multiple heat transfer problems for a 30 mm composite part subjected to similar air temperature cycles but with different BCs using a trained PINN.*

Figure 8 illustrates the ability of this neural network to accurately predict the temperature for a range of heat transfer coefficient combinations. The scenario illustrated in Figure 8 is that of a 30 mm composite bar initially at 0°C being heated by air that ramps linearly from 0°C to 50°C at a rate of 5°C/min, and that then holds constant at 50°C. The individual plots in Figure 8 show the temperature across the length of the bar, at time = 15 minutes, for selected heat transfer coefficient combinations. PINN results were validated by comparison with FE results, which performed separately for each of the scenarios. If the same problem needed to be solved for different combinations of heat transfer coefficients, a new finite element analysis would need to be run for each such combination. The PINN that includes the heat transfer coefficient as inputs, however, is already trained to make predictions for any combination of heat transfer coefficients. Given that the neural network is already trained, the computational costs of making predictions for new combinations of heat transfer coefficients is very low.

5. **Extension to 2D and Beyond**

The expansion of the PINN to two dimensions required only the addition of one input ($y$-dimension), the creation of another pre-layer for the $y$ input, the adjustment of the error calculation with respect to the PDE to include a term for the second derivative of the PINN prediction with respect to $y$, and the adjustment of the error calculation for the heat transfer boundary conditions. The addition of another input caused the model to require more total data points to train, but since the training data points were generated randomly for each epoch, this only resulted in the model requiring more epochs to train to an acceptable loss. Once the model was trained, the temperature at any point could be predicted by evaluating the model at the desired $x$, $y$, $t$, $h_1$, and $h_2$. Figure 9 illustrates this capability by showing a heat map of the temperature throughout a 60 mm by 20 mm composite part, with heat transfer coefficients of 100 W/m²K on each the $x = 0$ and $y = 0$ boundaries, with the same air temperature profile illustrated in Figure 5 (a). The three individual illustrations within Figure 9 show the temperature distributions at 5 minutes, 10 minutes and 15 minutes into the problem. The results were validated by comparison with FE results.

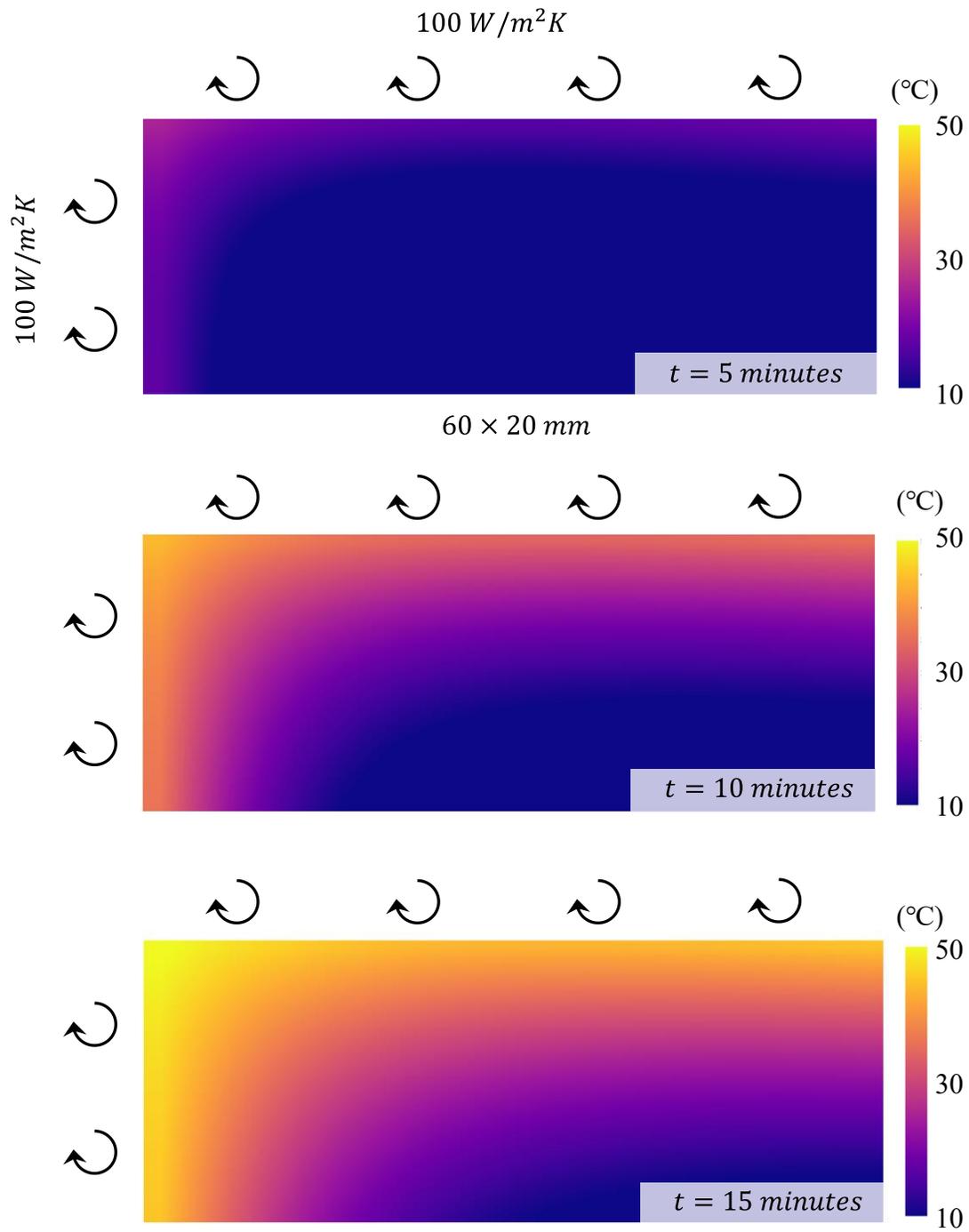

*Figure 9: Predictions of heat transfer in a 2D case using the trained PINN. Results are shown after 5, 10 and 15 minutes subjected to an air temperature in Figure 5 (a).*

### 5.1. *Further Extension*

The expansion of the PINN to three dimensions would require the addition of one input ($z$-dimension), another pre-layer for the $z$ input, the adjustment of the error calculation with respect to the PDE to include a term for the second derivative of the NN prediction with respect to $z$, and the establishment of error terms for any additional heat transfer boundary conditions. Additional heat transfer boundary conditions may also require the addition of heat transfer inputs to the PINN. The addition of more variables will further increase the batch size and number of epochs required to train the model to an acceptable loss.

The neural network may also be modified to include the material properties as inputs. For the one-dimensional problem, it is sufficient to use the thermal diffusivity $\alpha$ and the thermal conductivity $k$ as inputs, but to solve the problem in 2- and 3- dimensions the specific heat capacity ($\rho c_p$) and thermal conductivity vector ($\vec{k} = (k_x, k_y, k_z)$) should be used. Adding two or three more inputs to the 2D or 3D neural network further increases the batch size and number of epochs that would be required to train the model to an acceptable loss.

The air temperature profile used to train the model was explicitly captured within the training algorithm, necessitating a new model to be trained for each air temperature profile. The model could be expanded to include a set of parameters that define a family of air temperature profiles, such as one parameter for the initial temperature, one parameter for the temperature ramp rate, one parameter for the hold temperature, one parameter for the hold duration, another parameter for a second temperature ramp rate and a final parameter for a final temperature could be used to describe a family of air temperature profiles. This specific parameterization would require the addition of five input variables to the model.

Each of these potential extensions as additional input variables to the model and thus increases the batch size and number of epochs required to train the model to an acceptable loss. Simultaneously implementing multiple extensions may push the computational power required for training beyond the limits of most individual computers.

## 6. Conclusions

In this study, a physics-informed machine learning approach was developed to solve the heat transfer PDE with convective BCs. This was based on training a neural network using a loss function defined to simultaneously satisfies the PDE, BCs and IC. In addition, feature engineering was used to define NN features based on the heat transfer theory. For accurate training, an adaptive normalizing scheme was developed and implemented to address the difference in magnitudes of loss terms. In addition, the selection of training points in each batch was designed to increase the density of training points around the discontinuities of loss function and inputs.

The predictions of the trained PINN were validated in several 1D and 2D heat transfer cases by comparison with FE results. In addition, performance and accuracy of the PINN was compared with those of a NN without feature engineering. It was shown that while both NN and PINN match the FE results within the training zone, only the PINN with engineered features can capture the physics of the problem to make accurate predictions beyond the training zone.

Developing PINNs to solve heat transfer PDE and other similar PDEs offers tremendous advantages over traditional approaches using FE simulations. Once trained, a PINN can be used for near real-time simulation capability of problems with any given BC. For manufacturing problems, considering uncertainties in the process including unknown BCs, such an approach allows for quick evaluation of the problem during processing to develop feedback loops, realizing the Industry 4.0 concept of active manufacturing control based on sensor data.